\documentclass[sigconf]{acmart} 

\usepackage{booktabs} 
\usepackage{hyperref}
\usepackage{amsmath,amssymb,mathtools}
\usepackage{xr}
\usepackage{float}
\usepackage{algorithm}
\usepackage[noend]{algpseudocode}

\begin{document}
\copyrightyear{2018} 
\acmYear{2018} 
\setcopyright{acmcopyright}
\acmConference[GECCO '18]{Genetic and Evolutionary Computation Conference}{July 15--19, 2018}{Kyoto, Japan}
\acmBooktitle{GECCO '18: Genetic and Evolutionary Computation Conference, July 15--19, 2018, Kyoto, Japan}
\acmPrice{15.00}
\acmDOI{10.1145/3205455.3205615}
\acmISBN{978-1-4503-5618-3/18/07}
\title{Combating catastrophic forgetting with developmental compression.}


\author{Shawn L.E. Beaulieu}
\affiliation{%
  \institution{University of Vermont}
  \city{Burlington, VT, USA} 
}
\email{shawn.beaulieu@uvm.edu}

\author{Sam Kriegman}
\affiliation{%
  \institution{University of Vermont}
  \city{Burlington, VT, USA} 
}
\email{sam.kriegman@uvm.edu}

\author{Josh C. Bongard}
\affiliation{%
  \institution{University of Vermont}
  \city{Burlington, VT, USA}
}
\email{josh.bongard@uvm.edu}

\renewcommand{\shortauthors}{Beaulieu et al.}

\begin{abstract}


Generally intelligent agents exhibit successful behavior across problems in several settings.
Endemic in approaches to realize such intelligence in machines is catastrophic forgetting: sequential learning corrupts knowledge obtained earlier in the sequence, or tasks antagonistically compete for system resources. Methods for obviating catastrophic forgetting have sought to identify and preserve features of the system necessary to solve one problem when learning to solve another, or to enforce modularity such that minimally overlapping sub-functions contain task specific knowledge. While successful, 
both approaches scale poorly because they require larger architectures as the number of training instances grows, causing different parts of the system to specialize for separate subsets of the data.
Here we present a method for addressing catastrophic forgetting called developmental compression. It exploits the mild impacts of developmental mutations 
to lessen adverse changes to previously-evolved capabilities and `compresses'
specialized neural networks into a generalized one.
In the absence of domain knowledge, developmental compression produces systems that avoid overt specialization, alleviating the need to engineer a bespoke system for every task permutation and suggesting better scalability than existing approaches. We validate this method on a robot control problem and hope to extend this approach
to other machine learning domains in the future.
\end{abstract}

%
%



\maketitle

\section{Introduction}
\label{sec:introduction}

The method introduced here resists catastrophic
forgetting on a simple task by exploiting the evolution of development.
We thus discuss below relevant work from these
two domains.

\subsection{Catastrophic forgetting.}

It has been known since the early days of 
machine learning research that catastrophic interference 
\cite{mccloskey1989catastrophic}, now more commonly
known as catastrophic forgetting 
\cite{french1999catastrophic, goodfellow2013empirical}, 
is a major
challenge to training neural networks effectively.
Even for the most common forms of network training
such as the backpropagation of error, there is
no guarantee that reducing the network's error
on the current training sample does not increase
error on the other samples.

For these reasons, much effort has been expended
to address this problem. One family of solutions
involves constructing modular networks
\cite{lipson2002origin, ellefsen2015neural,
clune2013evolutionary, espinosa2010specialization,
kashtan2005spontaneous, sabour2017dynamic}
in which different modules deal with
different subsets of the training set. In such
networks, changes to one module may result in
improved performance for the training subset
associated with that module without disrupting
performance on other subsets. Such modularity
has indeed been demonstrated to minimize
catastrophic forgetting \cite{ellefsen2014neural, 
rusu2016progressive, 
lee2016dual, 
rusu2016progressive, 
lee2016dual, fernando2017pathnet}. 
Related to this concept of modularity are networks
in which some subsets of the network that have a
large impact on the current training set are
made less resistant to change during subsequent
training
\cite{kirkpatrick2017overcoming, velez2017diffusion}. 
The remaining parts of the network,
which remain adaptive, are thus able to fit to new training instances without disrupting
behavior on previous instances.

The drawback of
these approaches however is that network size tends
to increase with the amount of training data,
because new modules must be implicitly
or explicitly added for new training data. In the work
presented here, we introduce a method for combating
catastrophic forgetting by drawing on the idea of
compression: the size of the learner expands when
new training instances are encountered, but the learner
is then compressed back to its original size in a gradual manner through evolutionary canalization.

Besides modularity, another guard against catastrophic
forgetting is to reduce the magnitude of
behavioral impact after some change is made during
training. The intuition here is that small changes
to network behavior may increase the likelihood
of local improvements for new training instances
while minimizing or nullifying performance decreases
on the previous training set. 

In evolutionary methods, one way of reducing
behavioral impacts is to dynamically tune, during
evolution, mutation rates 
\cite{dang2016self}
and/or crossover events
\cite{teo2016fixed}. A recent approach demonstrated
for neuroevolution  is to dynamically tune
individual synaptic weights proportionally to their
impact on the network's behavior \cite{lehman2017safe}.
Similarly,
in the genetic programming community, semantic
variation operators have been reported
\cite{Vanneschi_2014_A-Survey, Castelli_2014_Semantic,
pawlak2015semantic, szubert2016semantic}.
These operators take into account the semantics of subtrees 
or individual tree nodes, and attempt to replace them with 
new genetic material that exhibits similar semantics.

Here, we introduce a new method that exploits 
development---the fact that some agents may change their internal structure over their lifetimes---to reduce the behavioral impact of evolutionary perturbations. This reduced behavioral impact, which results in late onset behavioral change, then facilitates the compression of networks specialized in different training subsets back down to one network. Compression is mediated by conditions that allow for the Baldwin effect, in which beneficial behaviors that manifest late in life occur increasingly earlier in successive generations. 

\subsection{The evolution of development.}

The evolution of development, or `evo-devo' has received
increasing scrutiny in AI and robotics research.
\cite{hinton1987learning} showed that combining learning
and evolution can smooth the fitness landscape and thus
facilitate evolution. Since learning can be considered
a form of development
\cite{kouvaris2017evolution}, that experiment can thus be
considered the initial investigation into how combining
evolutionary change with lifetime change can facilitate
evolution. 
In evolutionary robotics, several evo-devo
approaches were investigated
\cite{Dellaert94, miller2003evolving}, 
some of which involved considering
morphological change in the robot over its lifetime
\cite{eggenberger1997creation, Bongard03, Bongard11}.
This work was followed by the introduction of compositional
pattern producing networks (CPPNs; \cite{stanley2007compositional}), 
which are described
as capturing one important aspect of development: its ability
to bias evolutionary search toward solutions with regular
internal structure.

Despite their clear ability to increase evolvability in 
many systems, CPPNs do not model one important aspect of
biological development: change over the lifetime of the
learner. The very fact that agents may change over time
provides a unique opportunity to increase their evolvability.
If a mutation alters the developmental program of an agent,
the behavioral impact of that mutation may not manifest
until later in the descendant agent's lifetime. If the
fitness of an agent is integrated over its lifetime, it
follows that the overall behavioral impact of a developmental
mutation is likely to be less than that of a non-developmental
mutation. This follows because a non-developmental mutation takes immediate effect at the beginning of an agent's lifetime.

Last, the likelihood that a mutation
is beneficial is inversely proportional to its magnitude of behavioral
change. Thus, a developmental mutation which has a late
onset effect, and so less behavioral impact, is more likely to be beneficial than a non-developmental
mutation. In effect, one can increase the likelihood of
beneficial mutations, and thus overall evolvability,
simply by including developmental change in populations of 
evolving agents. That development unlocks broader and more nuanced ranges of 
behavioral change
compared to non-developmental programs has recently been documented
\cite{kriegman2017minimal, kriegman2017morphological}.

In the next section we describe how evo-devo methods can 
be used to guard against catastrophic forgetting by gently mutating learners in such a way that improves their ability on one training instance
without damaging performance
on other training instances.

\begin{figure}[t]
\centering
\includegraphics[width=\linewidth]{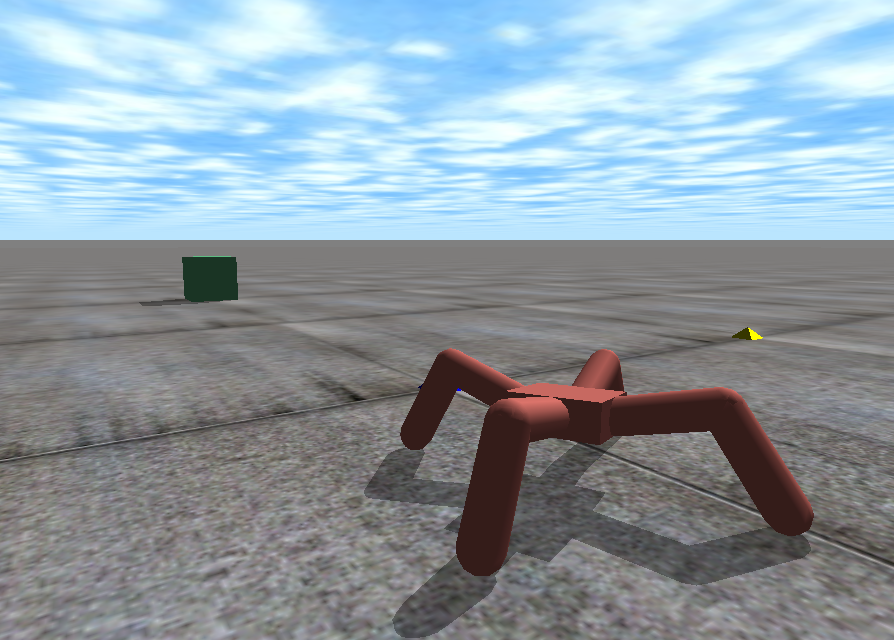}
\caption{\label{fig:robot}
A screen shot of environment $\boldmath{A}$, in which a light-emitting block is placed 30 body lengths \textit{in front of} the robot's starting position.
In environment $\boldmath{B}$, the block is placed 30 body lengths \textit{behind} the robot.
The robot is to perform phototaxis with an on-board sensor that detects light intensity according to the inverse square law.
}
\end{figure}
\section{Methods}
\label{sec:methods}

We chose to test developmental compression on a robot control problem for three
reasons. First, the continuous control of legged robots is a notoriously difficult
machine learning problem. Second, by dint of the problem's inherent difficulty,
if a robot is exposed to multiple disparate environments, catastrophic forgetting
is likely to occur. That is, improvements to the robot's ability in one 
training environment
will likely disrupt previously-evolved capabilities in other environments. Third, a robot's behavior at one time step has a cascading effect on its
behavior in future time steps; so, if fitness is integrated over multiple time steps, and developmental change
in control policy is allowed, developmental mutations
that manifest late in the life of individuals will impact overall
behavior less. This is in contrast with non-developmental mutations which, by definition,
take effect at the first time step of a robot's evaluation. This difference is exploited by developmental compression, as explained in \ref{sec:methods:DC}.

\subsection{The robot.}
\label{sec:methods:robot}

The robot and its environment were simulated using 
Pyrosim\footnote{\href{https://ccappelle.github.io/pyrosim/}{ccappelle.github.io/pyrosim}}
\cite{kriegman2017simulating}, an open source Python wrapper built atop the Open Dynamics Engine (ODE) 
physics engine \cite{smith2008open} (Fig. \ref{fig:robot}).

The architecture, or morphology, of the robot is generic by design. It was not selected to optimally solve the task it was given and, indeed, was chosen before the task environment was fully specified. It is characterized by a rectangular abdomen, attached to which are four legs, each composed of an upper and lower cylindrical object. The
knee and the hip joint of each leg contain a rotational hinge joint with one-degree-of-freedom. Each hinge joint can flex inward or extend outward by up to 90 degrees away
from its initial angle. The orientations of the hip and knee joints are set such that
each leg moves within the plane defined by its upper and lower leg components. 

Inside each lower leg is a touch sensor neuron, which at every time step detects when the lower leg to which it belongs makes contact with the environment.
It takes on either $-1$ (no contact) or $+1$ (contact).
Apart from the four touch sensors, a light sensor (whose output is a floating-point number) is embedded in the abdomen of the robot. Due to the inverse square law of light propagation, the light sensor is set
to $\ell = 1 / d^2$, where $d$ is the Euclidean distance between the light sensor and the light source. Motor neurons innervate each of the eight joints of the robot and enable it to move. 

\subsection{The controller.}
\label{sec:methods:controller}

In total, there are five sensor neurons fully connected to eight motor neurons such that each sensor neuron feeds into every motor neuron. No hidden neurons were employed.

How the set of sensor neurons connects to, and communicates with, the motor neurons is determined by the robot's neural controller. The edges of the network are represented by a weighted adjacency matrix that is optimized by a direct encoding evolutionary algorithm. Synaptic weights are constrained to $[-1,+1]$.

Motor neurons are updated according to 
\begin{equation}
m_i^{(t)} = \tanh\ \left[ m_i^{(t-1)} + \tau_i \sum_{j=1}^{J=5} w_{ji} s_j^{(t)} \right]
\end{equation}
where 
$m_i^{(t)}$ denotes the value of the $i$th motor neuron at the current
time step, 
$m_i^{(t-1)}$ is a momentum term that guards against `jitter' 
(high-speed and continuous reversals in the angular velocity of a joint),
$\tau_i$ is a time constant that can strengthen or weaken the influence of sensation
on the $i$th motor neuron relative to its momentum,
and $w_{ji}$ is the weight of the synapse connecting the $j$th sensor neuron to the
$i$th motor neuron. In order to ensure that random controllers produce diverse
yet not overly-energetic motion, all $\tau_i$ were set 0.3 via empirical
investigation.

\subsection{The task environment.}
\label{sec:methods:env}

When the simulation begins the robot is placed at a location that is identically equal to the origin of a smooth two-dimensional plane. 
Two tasks are presented sequentially to the robot. For the first, a light-emitting box is placed 30 body lengths away from the origin in the $positive$ y-direction (that is, $into$ the screen); for the second, the box is placed 30 body lengths from the origin in the $negative$ y-direction. Success is defined as the ability to walk toward the box in both environments in the alloted time (1000 ts).

\subsection{The fitness function.}
\label{sec:methods:fitfn}

As in the real world, the decay of a light signal's strength is proportional to the inverse squared distance from the source location. Consequently, fitness is taken to be the mean light sensor value experienced by the robot during the simulation. Integrating over time imposes a soft constraint on long peregrinations that obtain good performance only at the end of the evaluation. This selects for behavior of high proficiency throughout the life of an individual. Fitness increases both with proximity to the light source and with increasingly economical behavior that results in quicker travel.

Formally, fitness is defined as:
\begin{eqnarray}
\label{eq:fitfn}
f & = & \sum_{e=1}^{E} \frac{1}{T} \sum_{t=0}^{T} p_{t}^{(e)}
\end{eqnarray}
where $p_{t}^{(e)}$ denotes light sensor's value in environment $e$ at time $t$

In the experimental treatment each individual is evaluated 4 times, whereas the control treatment has only two evaluations per individual. To ensure equality of resources, the number of generations in the experimental treatment is half the number used in the control treatment.

\subsection{The compression algorithm.}
\label{sec:methods:DC}

\begin{figure}
\centerline{\includegraphics[width=\linewidth]{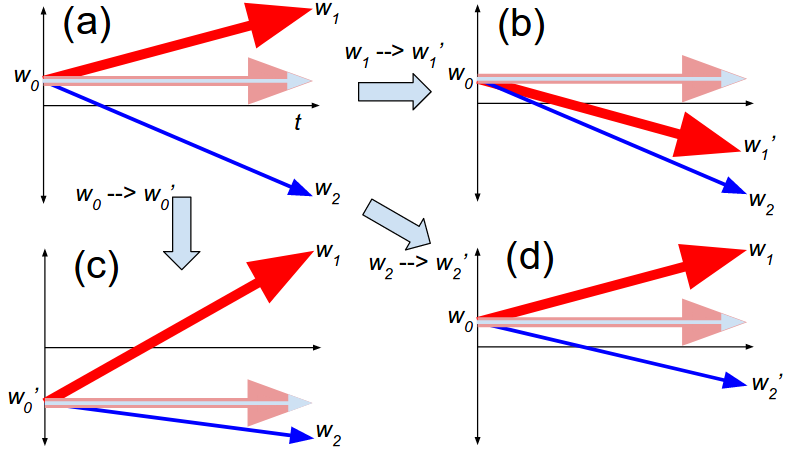}}
\caption{\label{fig:concept}
A conceptual visualization of the developmental compression algorithm on a single synaptic
weight. 
(a) Evolution tries to locate a compressed representation of target weights $w_{1}$ and $w_{2}$. (b) Mutation affects $w_{1}$. (c) Mutation affects base $w_{0}$. (d) Mutation affects $w_{2}$.
Horizontal axes represent developmental time.}
\end{figure}

The intuition for developmental compression is that a given neural controller contains a number of
sub-controllers which evolve to specialized, successful behavior in separate training environments.
Then, mild developmental mutations gradually `compress' the specialized sub-controllers into a single base controller, $W_{0}$.

More specifically, in two task environments, the networks we seek to compress are those that comprise an individual's genetic tensor $(5\times8\times3)$, represented by three distinct weight matrices $(5\times8)$. One such matrix serves as the base controller, while the other two are the target specialist controllers toward which the base linearly develops in the corresponding task environment. Concretely, the zeroth sheet of the tensor, $W_{D}^{(0)}$, is the base controller while $W_{D}^{(e)}$ is the target controller for the $e^{th}$ environment. 

Development is the process by which compression is enforced. It works by moving the weights of the base state toward a given target state as the simulation proceeds. Under the developmental treatment for the first task, the robot begins the simulation with controller $W_{D}^{(0)}$ and ends the simulation with controller $W_{D}^{(1)}$ under a development schedule that's linear in time.

Generically, for a target state belonging to task environment $e$, every element of $W_{D}^{(0)}$ moves toward $W_{D}^{(e)}$ by:
\begin{eqnarray}
w_{ji}^{(0)} \rightarrow w_{ji}^{(e)} = \alpha_{t}w_{ji}^{(0)} + (1-\alpha_{t})w_{ji}^{(e)}
\end{eqnarray}
where $\alpha$ is a weight ranging from 1 to 0 that linearly decays with $t$ that moves
from 0 to $T-1$ over $T$ time steps in the evaluation. 
Thus, the weight of synapse $w_{ji}$ at time $t$ becomes 
\begin{eqnarray}
w_{jit} = \frac{(T-1-t) w_{ji}^{(0)} + t w_{ji}^{(e)} }{T-1}
\end{eqnarray}
As the evaluation proceeds, $\alpha$ will dampen the base element and amplify the target element (line 32 in Algorithm \ref{alg:DC}).

For every task environment, a single agent is evaluated twice: once developmentally and once non-developmentally
(lines 23 and 24 in Algorithm \ref{alg:DC}, respectively). 
The fitness they obtain in each treatment is then added to their total fitness. 

For mutation, a single synaptic weight is chosen at random within the 
$ S \times M \times (e + 1) $ weight matrix $W$: $w_{jik}$.
The new weight of $w_{jik}$ is sampled from a Gaussian distribution whose mean and standard deviation depend on the selected synapse's prior value:
\begin{equation}
\label{eq:mutation}
w_{jik}^{(\text{new})} \sim \mathcal{N}\left(\,\mu=w_{jik}^{(\text{old})}, \;
\sigma
=|\,w_{jik}^{(\text{old})}\,|\,\right)
\end{equation}
If a mutation carries a weight above 1, it is set to 1; if a mutation carries
a weight below -1, it is set to -1 (Fig.\ref{fig:concept}). 

\subsection{The control algorithm.}
\label{sec:methods:control}

The control algorithm optimizes performance by summing fitness scores of non-developmental agents across all task environments. That is, individuals are evaluated once in each environment and the fitness they obtain is added to their respective total fitness, $F_{p}$, which is reported once all environments in $E$ have been encountered. 

Mutation here mirrors Equation \ref{eq:mutation}, 
except that one synaptic weight is chosen at random from the $5 \times 8$ weight matrix, not the genetic tensor.

\subsection{Random search algorithm.}
\label{sec:methods:random}

As a sanity check, the compression and control algorithms were compared with a random search algorithm, in which heredity is discarded. In other words, random search is the control treatment with reproduction removed. However, the ways in which individuals are mutated and evaluated are unchanged.

The total number of robot evaluations across all three algorithms is equalized to ensure
a fair comparison.
\setlength{\textfloatsep}{14pt} 

\begin{algorithm}
\caption{Developmental Compression vs. Control \label{alg:DC}}
\begin{algorithmic}[1]
\State $E \gets $ The Task Environments $(1 \times \text{number of envs})$
\State $W_{S} \gets $ Static Genome $(5 \times 8 \times 1)$
\State $W_{D} \gets $ Developmental Genome $(5 \times 8 \times \text{number of envs})$
\State $G \gets $ Number of Generations

\State
\If{controlTreatment}
	\For{g=1:$G$}
		\State $\textsc{Control}\,(\, W_{S} ,\, E \,)$
	\EndFor
\EndIf
\If{experimentalTreatment}
	\For{g=1:$G$/2}  
		\State $\textsc{Developmental\_Compression}\,(\, W_{D} ,\, E \,)$
	\EndFor
\EndIf
\State
\Procedure{Control}{$\,W_{S}\,,\, E\,$} 

    \For{each individual in the population}
    \State 
    \texttt{Fitness}
    = 0 
		\For{$e \; \textbf{in} \;  E$}
			\State 
            \texttt{Fitness}
            += \textsc{Sim}\,$\left(\,e,\, 	\text{base}=W_{S},\,\text{target}=W_{S}\, \right)$
		\EndFor
	\EndFor    
\EndProcedure
\State
\Procedure{Developmental\_Compression}{$\,W_{D}\,,\, E\,$} 
    \For{each individual in the population}
    \State 
    \texttt{Fitness}
    = 0
		\For{$e \; \textbf{in} \;  E$}
			\State 
            \texttt{Fitness}
            += \textsc{Sim}\,$(\,e,\, 		\text{base}=W_{D}^{(0)}, \, \text{target}=W_{D}^{(e)}\;)$
			\State 
            \texttt{Fitness}
            += \textsc{Sim}\,$(\,e,\, 		\text{base}=W_{D}^{(0)},\,  \text{target}=W_{D}^{(0)}\;)$
		\EndFor
	\EndFor
\EndProcedure
\State
\Procedure{Sim}{$e$, base, target}
	\State $p_e$ = 0 \Comment{performance (light intensity)}
	\For{$t$=1:T} \Comment{time steps}
    	\For{$s$=1:S} \Comment{sensor neurons}
        	\For{$m$=1:M} \Comment{motor neurons}
                \State
            	\State $w_{smt} = ( (T-1-t) w_{sm}^{(\text{base})} + \; t w_{sm}^{(\text{target})} ) / (T-1)$
            \EndFor
        \EndFor
        \State Perform in env $e$ for time step $t$ using $W_t$.
        \State $p^{(e)}$ += $p_{t}^{(e)}$
    \EndFor
    \State \textbf{return} $\,p_{t}^{(e)}$
\EndProcedure
\end{algorithmic}
\end{algorithm}

\externaldocument{methods}

\section{Results}
\label{sec:results}

We performed 100 independent runs of 
the experimental treatment  (750 generations per run, Sect. \ref{sec:methods:DC}), 
the control treatment  (1500 generations, Sect. \ref{sec:methods:control}), 
and random search (1500 generations, Sect. \ref{sec:methods:random}). For each run, the fittest individual across all generations was extracted: that is, the overall run champion, as defined by Equation \ref{eq:fitfn}. 

\begin{figure}[H]
\centerline{(a)\includegraphics[width=0.46\textwidth]{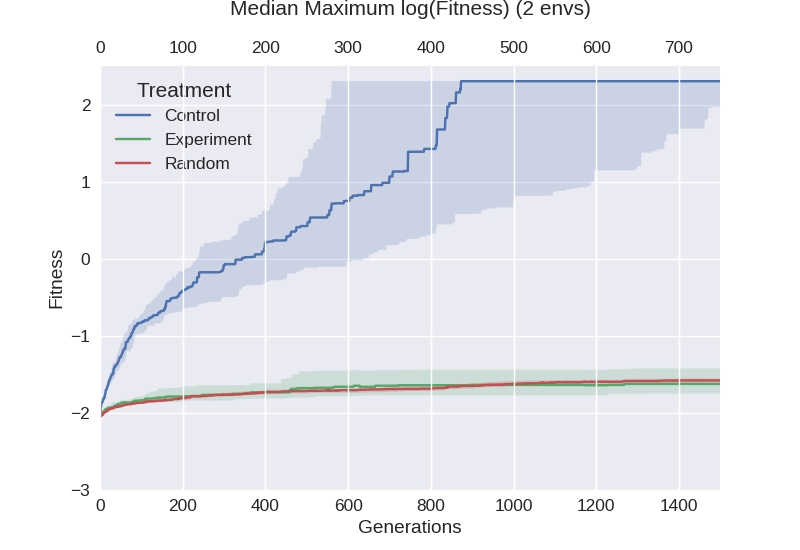}}
\centerline{(b)\includegraphics[width=0.46\textwidth]{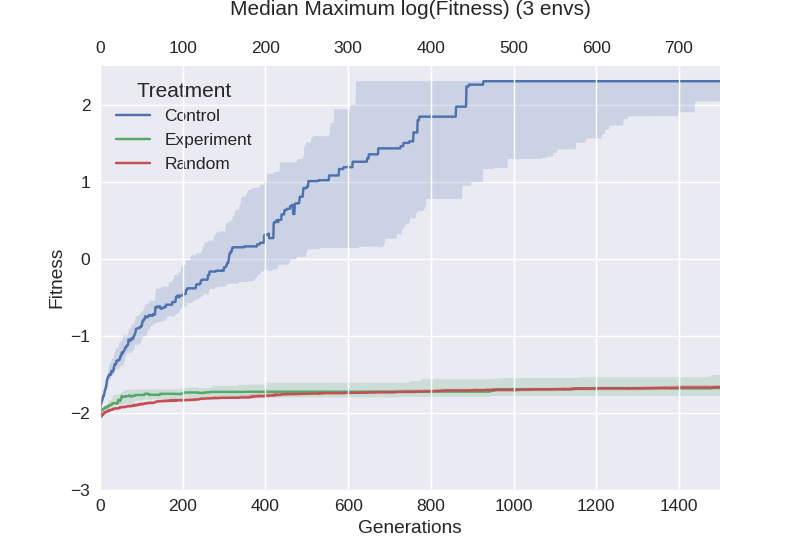}}
\centerline{(c)\includegraphics[width=0.46\textwidth]{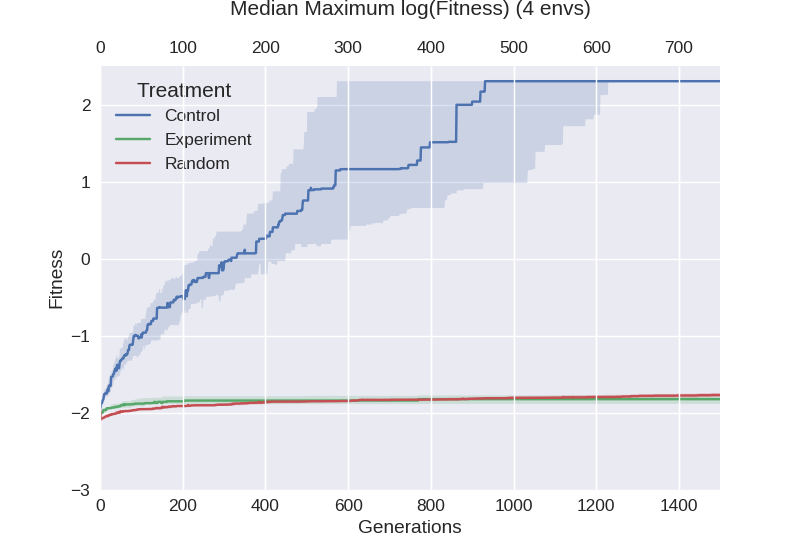}}
\caption{\label{fig:max_fit} Median Maximum Fitness. The median score across all runs for the maximum fitness value of generational run champions (95 percent confidence intervals). (A) two task environments (E=2); (B) three task environments (E=3); (C) four task environments (E=4). Y-axis is log-scale. For reference, with two environments the experimental treatment has median=0.19, std=0.202. The top axis indicates the generations seen by developmental compression}.
\end{figure}

\begin{figure}[H]
\centerline{(a)\includegraphics[width=0.46\textwidth]{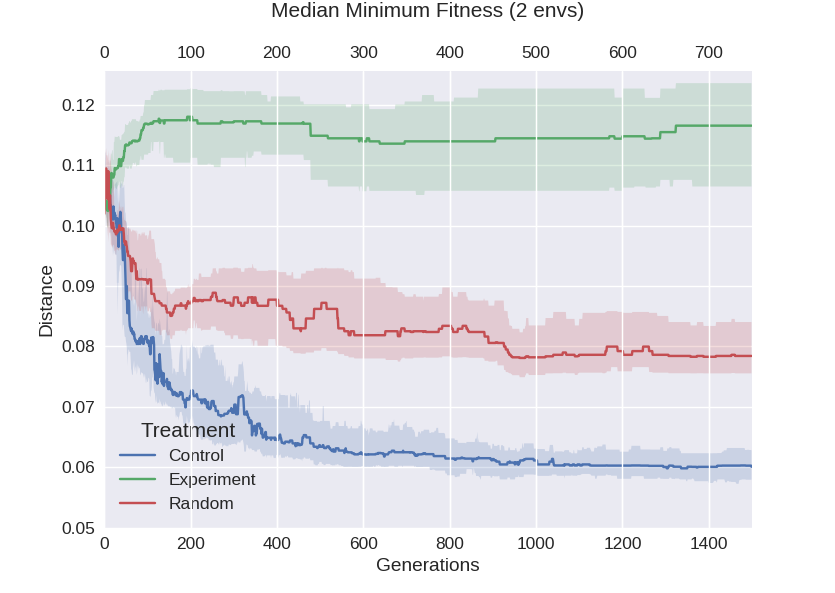}}
\centerline{(b)\includegraphics[width=0.46\textwidth]{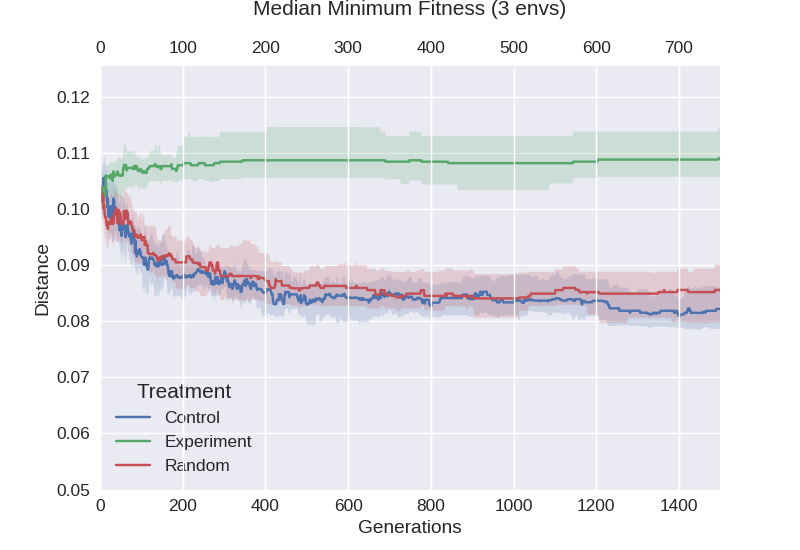}}
\centerline{(c)\includegraphics[width=0.46\textwidth]{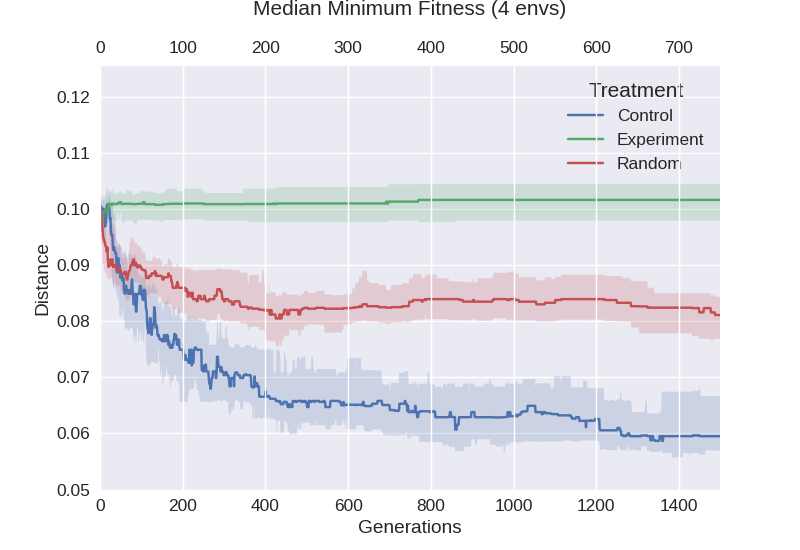}}
\caption{\label{fig:min_fit} Median Minimum Fitness. The median score across all runs for the minimum fitness value of generational run champions (95 percent confidence intervals). (A) two task environments (E=2); (B) three task environments (E=3); (C) four task environments (E=4). This figure highlights catastrophic forgetting in the random and control methods. Control is more adversely affected by forgetting due to overspecialization. The experimental method avoids the negative impact of task antagonism.
}
\end{figure}

The results presented in Figures 
\ref{fig:min_fit} and \ref{fig:max_fit} (generational champions) and 
\ref{table:results:fits} (overall champions) strictly pertain to non-developmental performance. In the DC treatment, run champions are those whose fitness is highest across all $2E$ evaluations, but we report only the non-developmental performance of those run champions 
(line 24 in Algorithm \ref{alg:DC})
for fairness of comparison. 

Fig. \ref{fig:max_fit}a reports the median fitness for the generational champions' best environment for all three treatments. 
This figure suggests that the control treatment is significantly outperforming both the experimental treatment and random search. 
However, Fig. \ref{fig:min_fit}a demonstrates that this conclusion is premature. This figure shows the median fitness of generational run champions in their worst environment.
Catastrophic interference is responsible for the discrepancy between the quantity being optimized by the evolutionary algorithm (mean fitness) and the quantity we implicitly seek to optimize (general proficiency). 

The ability of robots in the experimental treatment to continue improving in their worst environment is reported in Table \ref{table:results:fits}. 
For the case of two environments ($E = 2$), we assessed statistical significance for the median minimum fitness of each treatment's overall run champion with the Mann-Whitney U test and Bonferroni

\noindent
correction across three possible comparisons (compression vs. control, compression vs. random, random vs. control). 
We found
that developmental compression significantly out performed both
the control ($P < 0.0001$) and random treatments ($P < 0.0001$). Results for the median minimum fitnesses of overall run champions
are shown in table \ref{table:results:fits}. 

\begin{table}
\caption{\label{table:results:fits} Median minimum fitness of run champions.}
\vspace{-1em}
\begin{center}
\begin{tabular}{ | c | c | c | c | } \hline
					& Random 		& Control 		& Compression 	\\ 	\hline \hline 
E=2	& 0.0826 $\pm$ 0.0186 	& 0.0781 $\pm$  0.0372	& \textbf{0.1103} $\pm$ 0.0240 	\\ 	\hline
E=3	& 0.0870 $\pm$ 0.0188 	& 0.0847 $\pm$ 0.0229 	& \textbf{0.1094} $\pm$ 0.0188 	\\ 	\hline
E=4	& 0.0823 $\pm$ 0.0142 	& 0.0704 $\pm$ 0.023 	& \textbf{0.1016} $\pm$ 0.014 	\\ 	\hline
\end{tabular}
\end{center}
\end{table}

Minimum fitness is an important metric in this context because if it decreases over evolutionary time it reveals the degree to which an individual narrowly specializes to one task environment. This trend can also arise in the developmental treatment if the base genome evolves to be overly dependent on their scaffolds, sacrificing performance without them. A similar problem can afflict the non-developmental treatments in increasingly many environments, where a good strategy is to avoid costs to overall fitness by doing the same thing in every environment: stand still. 

\begin{figure}[t]
\centering
\includegraphics[width=\linewidth]{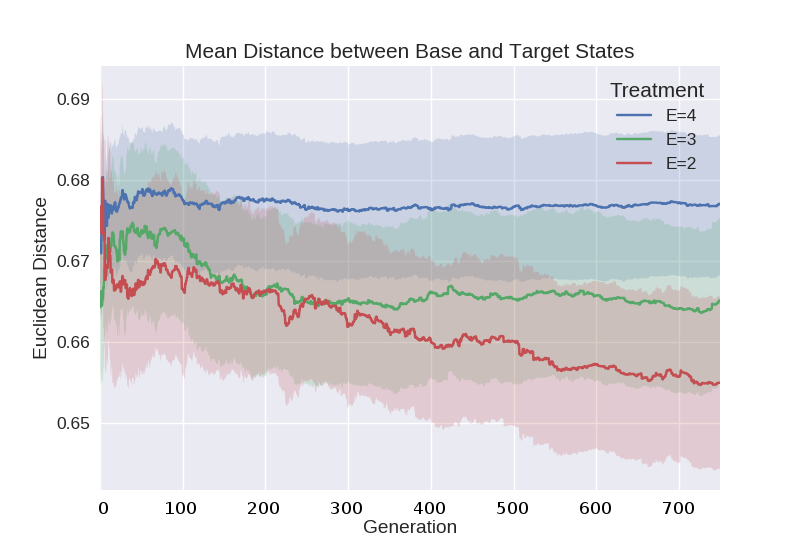}
\caption{\label{fig:compression} Compression over time. Mean distance during evolution between the base state and target states in the corresponding environments. Values computed using generational run champions across all runs for each treatment (E=2:4) with 95 percent confidence intervals.}
\end{figure}

The extent to which development compresses the weight matrices in an agent's controller is shown in \ref{fig:compression}. For every environment, the mean Euclidean distance between the base state and the target state for the generational run champions is computed over the course of evolution. The mean distance is then averaged over the number of environments and reported with a 95 percent confidence interval. At a glance, compressibility appears to correlate with the median minimum fitness plots in Figure \ref{fig:min_fit}, which suggests that DC's ability to avoid catastrophic forgetting depends on the quality of compression.   

By observing where and how individuals fail to achieve optimal performance where they otherwise should, we can glimpse into the logic under which they operate. For example, in the control, a high maximum fitness (Fig.\ref{fig:max_fit}) does not seem to correlate with the ability of phototaxis (Fig.\ref{fig:min_fit}), but rather with the ability to walk remarkably well in one direction.
These results suggest that developmental compression was partly able to overcome tempting local optima like specialization and inaction, while the control and random treatments were not.

\begin{figure}
\centering
\includegraphics[width=0.48\textwidth]{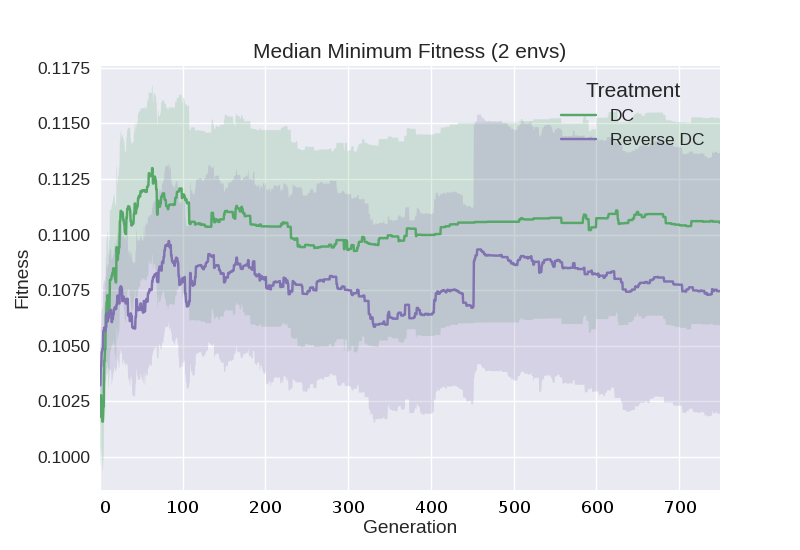}
\caption{\label{fig:reverse} DC vs reverse DC. Comparing the experimental treatment (green) and a reversed version (purple) with 95 percent confidence intervals. See Discussion for details.
}
\end{figure} 

\section{Discussion}
\label{sec:discussion}

Developmental compression appears to limit specialization, which afflicts both random search and
the naive approach of averaging non-developmental performance across all environments
(Fig. \ref{fig:min_fit}).
These latter two treatments likely specialize because they suffer from catastrophic
forgetting: seemingly redundant improvements in one's best environment recur
because such improvements outweigh the penalty to overall fitness incurred by improving one's worst environments.
Only robots within the developmental compression paradigm exhibit continued improvement in
all environments over evolutionary time 
(Figs. \ref{fig:max_fit}a and \ref{fig:min_fit}a).
This is may be ascribed to (at least) two factors.

First, mutations within developmental compression
tend to have more mild behavioral effects. This is illustrated in 
Fig. \ref{fig:concept}. 
A mutation that
strikes a feature in the base matrix, $w_{0}$, 
(Fig. \ref{fig:concept}c) 
impinges on behavior
in all $E$ environments, but this effect is attenuated as the base matrix develops toward its scaffolding. A mutation that strikes a target matrix 
(Fig. \ref{fig:concept}b) only modulates behavior in one of the $E$ environments. This is likely 
to increase the likelihood that the controller may improve in one environment without
adversely impacting its performance in the other $E-1$ environments.

The second reason that developmentally compressed controllers outperform those
from the random and control treatments at local task generalization is that evolution gradually ignores target scaffolding over evolutionary time. This can be seen in 
Fig. \ref{fig:compression} where
the distance between the base and target matrices decreases with time. This means that the three matrices within a controller are gradually
approaching one another, and the path traversed by development becomes shorter. This is canalization.

The success of developmental compression in this domain is even more surprising
given that its fitness landscape is twice as large as the control treatment.
However, work on extra-dimensional bypasses \cite{conrad1990geometry} has
shown that at high dimensions, if an optimization procedure can smooth the fitness space by bridging local optima in lower dimensional
spaces, evolvability can be increased. Thus developmental compression can be viewed
as a way to safely introduce extra-dimensional bypasses, which may militate against catastrophic
forgetting.

\subsection{Scalability.}

Despite its success in two environments, developmental compression weakened in its ability to obtain generalists as the problem was scaled up to three and four environments 
(Figs. 
\ref{fig:max_fit}b,c 
and 
\ref{fig:min_fit}b,c).
Indeed, as 
Fig. \ref{fig:compression} 
illustrates, DC also fails to compress 
increasingly many weight matrices as more problems are introduced.
At present it is difficult to say why this occurs. However, the most likely culprit
is a weak evolutionary algorithm and sparse neural controller. An extremely simple genetic algorithm was employed here---one that
doesn't attend to diversity, or other salient factors, and thus fails to obtain good and compressible scaffolds on complex problems.
In future work we plan to investigate how DC can be strengthened with better
evolutionary constraints, and be made to synergize with related algorithms.

Interestingly, the behavioral impacts of mutations become increasingly mild in DC as the number of task environments increases. For two environments ($E=2$), a mutation to a given target matrix impacts behavior in only one of environment; for three environments ($E=3$), a mutation to a target matrix impacts behavior in only one of three environments; and so on, with mutation becoming increasingly diluted.
While DC might dull the behavioral consequences of any single mutation, there's no guarantee that this will result in good generalists. Instead, this may slow the pace of
progress. In future work we will investigate how to address this by using adaptive mutation rates, rather than one that is fixed.

\subsection{Different forms of development.}

In this work, a simple process of development was employed: robots begin with a single controller and move toward $E$ controllers in the corresponding environments. As development proceeds, agents exhibit increasingly unique behaviors. However, different developmental trajectories---ones that aren't strictly linear---might be even better at avoiding catastrophic forgetting.

In this vein, we conducted another set of $100$ independent runs of developmental compression in each of two, three, and four environments. But in this case developmental interpolation was reversed: for developmental evaluations, $E$ base states developed linearly toward a $single$ target state (line 23 of Algorithm 1). Curiously, this reversed
variation did not appear to avoid catastrophic forgetting as well as the original experimental method
(Fig. \ref{fig:reverse}).
The suggests that the order of compression is important for success in this domain. How compression varies with different developmental programs will be the subject of future investigations. In 
Fig. \ref{fig:reverse} 
we only report performance for two environments, but the gap between DC and reverse DC widened more strikingly in three and four environments.

Another potential avenue for improvement is to change the entire process by which development occurs. In its current form, development is uniformly distributed across network parameters and is linear in time. But this need not be true. One could include time as an input to compositional
pattern producing networks \cite{stanley2007compositional} which would `paint' different, possibly nonlinear, 
developmental trajectories onto different synapses. One could also envision employing
NEAT \cite{stanley2002evolving} to compress base and target neural controllers with different cognitive architectures
into a single, non-developing yet generalist architecture.
Indeed, prior work has sought to marry network complexification and deep learning \cite{proGANs2017}, which may also be applicable to developmental compression.

\subsection{Synergies with other methods.}

There exist myriad ways to combat catastrophic forgetting, each motivated by specific problems and the obstacles inherent to them. We do not purport to show that developmental compression succeeds where prior methods fail, or that it is preferable in situations where catastrophic forgetting has already been reasonably tempered \cite{kirkpatrick2017overcoming}. Rather, we believe the results reported in this paper suggest developmental compression may be a potential mechanism for overcoming catastrophic forgetting---either independently or in concert with existing methods.

Although other techniques could be used to resist specialization, they often incur additional costs. For example, an objective that rewards generalists could be included in a multi-objective evolutionary algorithm. However, doing so would increase the dimensionality of the Pareto front, which is known to weaken optimization and requires bulwarking countermeasures \cite{deb2014evolutionary}, thus increasing algorithmic complexity. Another approach would be to formulate a better fitness
function. However, this would put us on a frictionless path toward fitness function engineering, which may be appropriate for narrow problems but is likely to produce systems that break outside of their training regime. One could also take the minimum fitness across $E$ to guard against specialization, but this will collapse gradients in the fitness landscape, because mutations that do not affect the worst fitness component aren't seen by search.

Unlike existing methods for combating catastrophic forgetting,
developmental compression is expressly designed to minimize the need for supervised intervention in the form of domain knowledge or explicit constraints on behavior. Rather than decide what features of the system are necessary for generalist capabilities, developmental compression lets evolution discover the best way to preserve knowledge through a development program that integrates information obtained over the life of the agent. Given the generality of this method, it is likely to synergize well with existing techniques.

Future implementations will seek to improve developmental compression and to reconcile the results reported here with the work being done in deep neuroevolution \cite{lehman2017safe}. The Atari-57 game suite and DMLab-30 environments are problems for which developmental compression could be scaled up, modified, and rigorously tested. Whether DC can be parallelized and used to compress populations of specialists across dissimilar tasks into a single master network remains to be seen. If such an ability can be realized, it would represent an evolutionary alternative to the multi-task learning paradigm currently being explored in reinforcement learning \cite{impala2018}.

\section{Conclusions}
\label{sec:conclusions}

Developmental compression is a novel technique for avoiding catastrophic forgetting. It works by relying on the smoothing effect of development, and how specialists
can thus be gradually and developmentally `compressed' into a generalist by canalization.
It could also work in domains beyond robotics provided two conditions are met. First, each training instance must be extended in time. That is, several forward passes are required to complete one evaluation. Such a condition is usually met by reinforcement learning tasks. Second, performance must be integrated over all time steps of the evaluation. In future work we plan to study the efficacy of developmental compression in machine learning domains outside robotics and in tandem with other advances in the artificial intelligence literature.

\section{Acknowledgments}

This work was supported by 
DARPA contract HR0011-18-2-0022. 
The computational resources provided by the 
UVM's 
Vermont Advanced Computing Core
(VACC)
are gratefully acknowledged.

\bibliographystyle{ACM-Reference-Format}
\bibliography{main}

\end{document}